\documentclass[lettersize,journal]{IEEEtran}
\usepackage{amsmath,amsfonts}
\usepackage{algorithmic}
\usepackage{algorithm}
\usepackage{array}
\usepackage[caption=false,font=normalsize,labelfont=sf,textfont=sf]{subfig}
\usepackage{textcomp}
\usepackage{stfloats}
\usepackage{url}
\usepackage{verbatim}
\usepackage{graphicx}
\usepackage{cite}

\usepackage{changepage,threeparttable} 
\usepackage[capitalize]{cleveref}
\usepackage[capitalize]{cleveref}
\crefname{section}{Sec.}{Secs.}
\Crefname{section}{Section}{Sections}
\crefname{table}{Tab.}{Tabs.}
\Crefname{table}{Table}{Tables}
\usepackage{multirow} 
\usepackage{soul}
\usepackage{xcolor} 
\usepackage{pifont}

\usepackage{amssymb}
\usepackage{booktabs}
\usepackage{float}
\usepackage{changepage,threeparttable}

\usepackage{multirow} 
\usepackage{soul}
\usepackage{pifont}
\newcommand{\cmark}{\ding{51}}%
\usepackage{changepage,threeparttable} 
\usepackage{float}

\usepackage{xspace}

\newcommand{\ie}{\emph{i.e.}\xspace}
\newcommand{\eg}{\emph{e.g.}\xspace}
\newcommand{\cf}{\emph{cf.}\xspace}

\newcommand{\tho}[1]{\textcolor{black}{#1}}

\usepackage{titlesec}
\titlespacing*{\section}{0pt}{0.5\baselineskip}{0.5\baselineskip}
\titlespacing*{\subsection}{0pt}{0.5\baselineskip}{0.5\baselineskip}

\begin{document}

\title{Accurate and Real-time 3D Pedestrian Detection Using an Efficient Attentive Pillar Network}

\author{Duy Tho Le, Hengcan Shi, Hamid Rezatofighi, Jianfei Cai\\
Department of Data Science and AI\\
Monash University, Australia\\
{\tt\small tho.le@monash.edu}
}



\maketitle

\begin{abstract}
Efficiently and accurately detecting people from 3D point cloud data is of great importance in many robotic and autonomous driving applications. This fundamental perception task is still very challenging due to \emph{(i)} significant deformations of human body pose and gesture over time and \emph{(ii)} point cloud sparsity and scarcity for pedestrian objects. Recent efficient 3D object detection approaches rely on pillar features. However, these pillar features do not carry sufficient expressive representations to deal with all the aforementioned challenges in detecting people. To address this shortcoming, we first introduce a stackable Pillar Aware Attention (PAA) module to enhance pillar feature extraction while suppressing noises in point clouds. By integrating multi-point-channel-pooling, point-wise, channel-wise, and task-aware attention into a simple module, representation capabilities of pillar features are boosted while only requiring little additional computational resources. We also present Mini-BiFPN, a small yet effective feature network that creates bidirectional information flow and multi-level cross-scale feature fusion to better integrate multi-resolution features. Our proposed framework, namely PiFeNet, has been evaluated on three popular large-scale datasets for 3D pedestrian Detection, \ie KITTI, JRDB, and nuScenes. It achieves state-of-the-art performance on KITTI Bird-eye-view (BEV) as well as JRDB, and competitive performance on nuScenes. Our approach is a real-time detector with 26 frame-per-second (FPS). The code for our PiFeNet is available at \normalfont{\emph{\textcolor{red}{\url{https://github.com/ldtho/PiFeNet}}}}.

\end{abstract}


\section{Introduction}

\begin{figure}[!h]
\begin{center}
\includegraphics[width=0.99\linewidth]{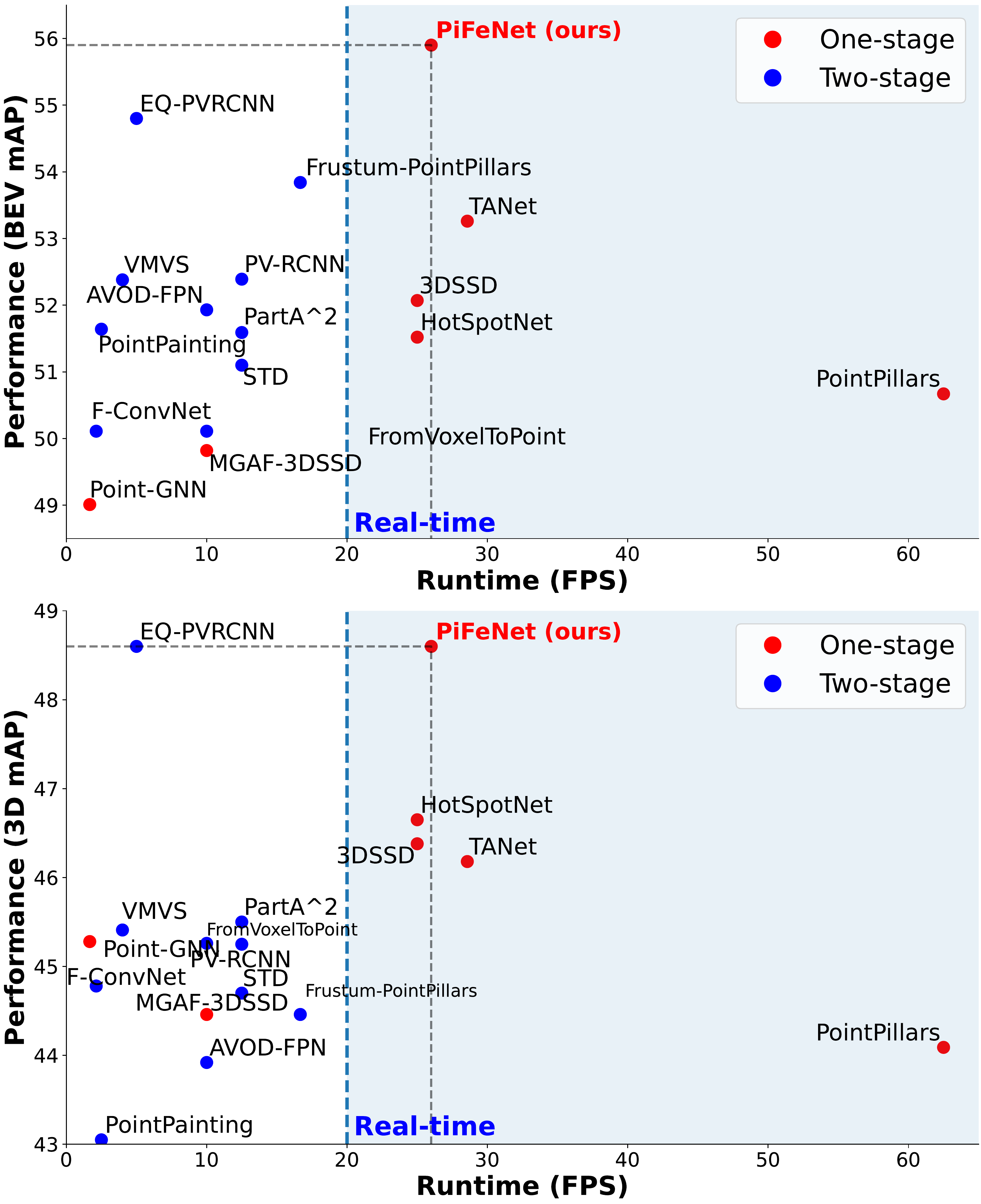}
\end{center}
\vspace{-1em}
   \caption{\tho{The performance of our PiFeNet on KITTI \cite{geiger2013vision} official test split on both pedestrian Birds-eye-view and 3D detection versus inference speed in FPS. Our approach achieved SOTA performance in detecting pedestrians. The comparison details are presented in \cref{table:pedestriancomparision}.}}
   \vspace{-2em}
\label{fig:comparision_plot}
\end{figure}

\IEEEPARstart{F}{ast} and reliable pedestrian detection in 3D world is a fundamental step in a perception system, conducive to many robotic and
autonomous driving applications such as human-robot interactions, service robots, and autonomous navigation in crowded human environments.
The recent 3D pillar-based object detectors, \eg~\cite{liu2020tanet, paigwar2021frustum, zhu2019class} integrate pillar features, a down-sampled representation of point clouds, into their frameworks to detect
objects. These approaches have shown auspicious results in detecting rigid and large objects such as vehicles. However, their effectiveness in detecting pedestrians is limited because of the two challenges.

\begin{figure}[tb]
\begin{center}
\includegraphics[width=0.95\linewidth]{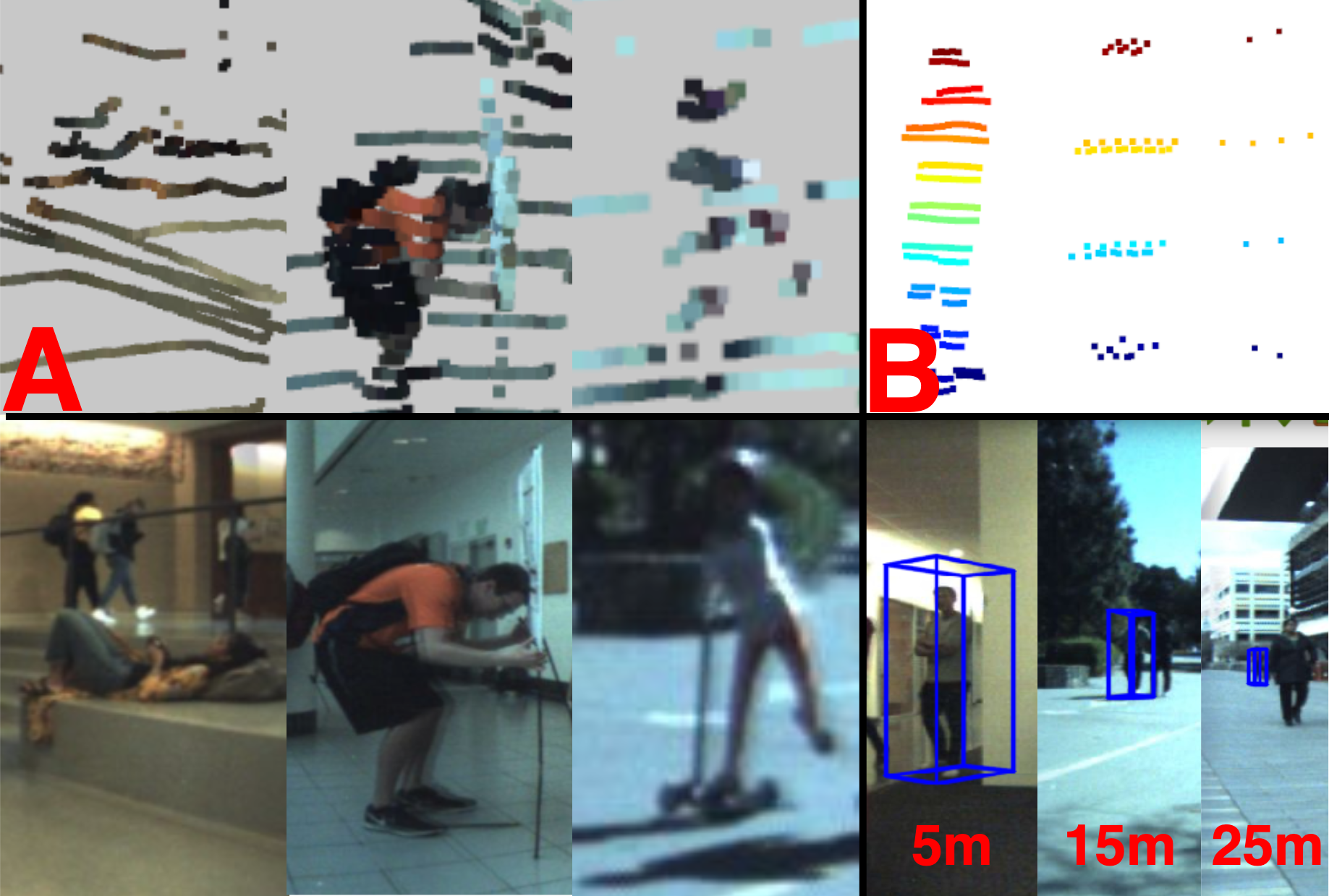}
\end{center}
\vspace{-1em}
   \caption{Point cloud density of pedestrian instances with challenging postures (A), and at increasing distances to the LiDAR sensor (B) captured in the JRDB\cite{martin2021jrdb} dataset (point clouds are zoomed and coloured for better visibility). Blue bounding boxes are instances of interest.}
   \vspace{-1em}
\label{fig:posedistviz}
\end{figure}

\textit{\textbf{Challenge 1}: Weak expressive capability of pillar features for pedestrians.} Pedestrians are not rigid bodies, hence they can take numerous poses (running, bowing, sitting, laying, etc.), as shown in~\cref{fig:posedistviz}A. Previous methods such as Pointpillars~\cite{lang2019pointpillars} struggle distinguishing pedestrians from poles and trees, due to 3D spatial information loss during quantisation process. They necessitate more representative pillar features to correctly distinguish pedestrians from other similarly-looking objects. To address this limitation, TANet~\cite{liu2020tanet} has incorporated a triple attention module to enhance pillar feature extraction. However, this approach only uses max pooling to capture context
features, suppressing a lot of information from the points inside the pillar, which are informative for better person detection and localisation.

To retain important information while generalising to various pedestrian representations, detectors require to adopt comprehensive attention mechanisms in pillar features. To this end, we propose a \textbf{Pillar Aware Attention (PAA) module} that performs attention on pillars utilising improved attention mechanisms. The PAA module includes multi-point-channel pooling, point-wise, channel-wise, and task-aware attention techniques to enhance pillar representations. Multi-point-channel pooling uses various pooling strategies to grasp context information from all points in a pillar. Point- and channel-wise attentions retain proper information during the feature extraction by suppressing redundant information. Task-aware attention switches the channels on or off based on their contributions to downstream tasks by selecting the most representative channels. We will show that PAA module is a key in boosting the performance of pedestrian detection. 

\textit{\textbf{Challenge 2}: Fewer pillars enclosing pedestrians after quantization as a result of their small occupancy and point sparsity in 3D point cloud.} In comparison with vehicles, human are relatively small, and thus occupy fewer pillars after the quantisation step. As demonstrated in \cref{fig:posedistviz}B, distant pedestrians are really small and hard to detect, due to their size and the sparsity of the sensor's laser beams. This becomes more challenging in crowded and/or cluttered areas, where people tend to stand or move in dense groups (\cf \cref{fig:JRDBviz}). 
Therefore, detecting pedestrians may require more than just extracting a naive pillar feature, \eg to enrich feature representations in both low- and high-resolution point clouds.

Thus, we introduce \textbf{Mini-BiFPN}, a lightweight feature network inspired by~\cite{tan2020efficientdet} that bidirectionally connects and fuses multi-resolution features.The learnable weights in Mini-BiFPN fusion gates can be adaptively adjusted based on the contribution of features at multiple resolutions. Also, the module is composed of three convolutional blocks with a minimal number of layers, making it lightweight and efficient.

Combining the aforementioned modules, we present \textbf{PiFeNet}, a novel end-to-end trainable 3D object detector capable of extracting important features in real-time, which benefits from the reduced inference latency of single-stage methods and the improved feature extraction mechanisms from attention. Our contributions are as follows:

\begin{enumerate}
    \item We introduce a stackable Pillar Aware Attention Module, modelling various attention on pillars for a better feature learning. 
    
    \item We propose Mini-BiFPN, a lightweight network that efficiently performs weighted multi-scale feature fusion and provides bidirectional information flows.
    
    \item We comprehensively evaluate the performance and generalisability of our framework on three large-scale 3D detection benchmarks, including KITTI \cite{geiger2013vision}, JRDB \cite{martin2021jrdb} and Nuscenes \cite{caesar2020nuscenes}. Compared with the published counterparts, PiFeNet achieves SOTA performance on KITTI (BEV) and JRDB, as well as competitive performance on nuScenes.
\end{enumerate}

\section{Related Work} \label{sec:relatedworks}
\begin{figure*}[!th]
\begin{center}
\includegraphics[width=\linewidth]{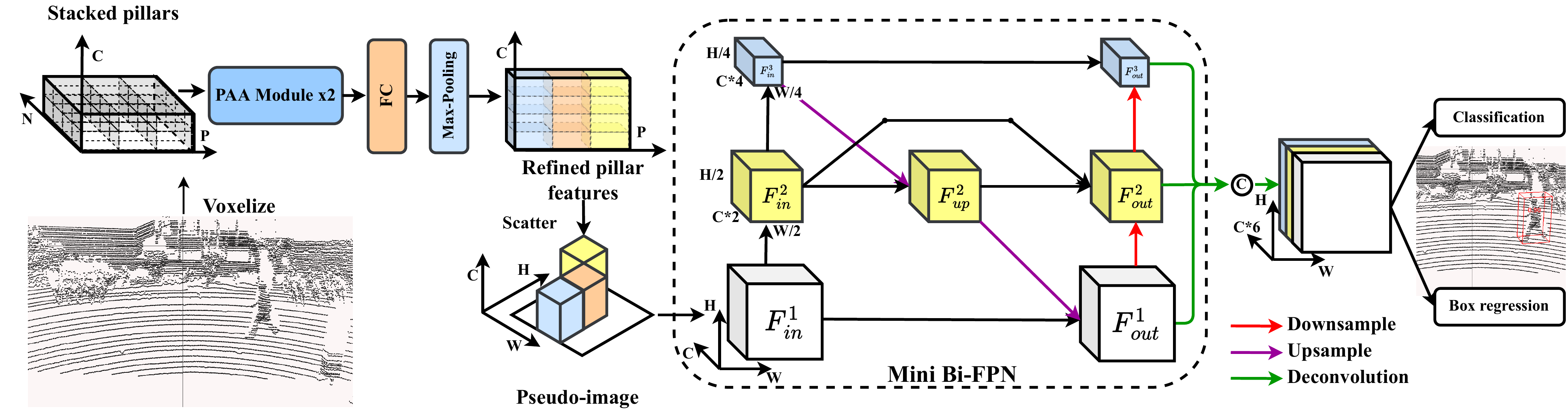}
\end{center}
\vspace{-1em}
   \caption{\tho{The PiFeNet's architecture. Input point clouds are first converted to pillars, and then passed through stacked PAA modules for feature extraction. The most expressive features are chosen and converted into a pseudo image, which serves as the input of Mini-BiFPN for generating 3D bounding boxes.}}
   \vspace{-1.5em}
\label{fig:PiFeNet}
\end{figure*}
\textbf{3D object detection from LiDAR point cloud.}
In general, 3D detectors for LiDAR point cloud can be classified into two types: one-stage and two-stage. Two-stage detectors \cite{shi2020pv,shi2019pointrcnn,chen2019fast,yoo20203d, xie2020pi, liang2019multi} first generate a series of anchors by the region proposal network (RPN), and then predict proposals based on these anchors. Many two-stage works \cite{shi2020pv,shi2019pointrcnn,chen2019fast} enhance RPN outputs, while others~\cite{yoo20203d, xie2020pi, liang2019multi} refine point cloud representations by multi-modal feature fusion from RGB images and 3D point clouds. \tho{Recently, Transformer-based two-stage approaches EQ-PVRCNN\cite{EQPVRCNN} and CasA\cite{wu2022casa} leverage the Query-Key mechanism to refine intermediate representations/proposals, serving as inputs to the latter stage}.

Although two-stage approaches provide more accurate predictions than one-stage detectors, they are significantly slower than one-stage counterparts due to the high computation overhead. One-stage approaches make a trade off between detection speed and accuracy. The recent progress on one-stage detectors \cite{yang20203dssd,zheng2021se,he2020structure} have been shown a very competitive performance compared to the two-stage approaches. They apply fully convolutional network on modified point cloud representations such as pillars~\cite{lang2019pointpillars} and voxels~\cite{zhou2018voxelnet, yan2018second}, to directly regress proposals. 
SA-SSD~\cite{he2020structure} further transforms convolutional feature maps to point-level representations to capture more structure-aware features. 3D-SSD \cite{yang20203dssd} proposes a real-time network by eliminating upsampling layers and the refining stage. CIA-SSD\cite{zheng2020cia} uses one fusion gate to link multi-scale features, while our Mini-BiFBN module presents four fusion gates.
Most recently, SE-SSD \cite{zheng2021se} proposes a shape-aware augmentation method to train a pair of teacher-student SSDs, which is encouraged to infer entire object forms.

These 3D detectors achieve great performance in detecting general object categories. Nevertheless, they do not perform well in detecting pedestrians due to aforementioned challenges. TANet~\cite{liu2020tanet} proposes a triple attention module, which can better extract discriminating features for pedestrians by attention mechanisms. However, this technique only considers the maximum values of the pillars' channels as encoded representations, and these maximum values do not reflect rich representations of all the points inside that pillar, which is essential for precise object detection and localisation. 
In this paper, we address this weakness by proposing PAA module with multiple attention mechanisms and improved pooling strategies.


\textbf{Attention mechanisms in object detection.}
Attention mechanisms have been shown to be advantageous for a variety of computer vision applications, including image classification \cite{he2020structure, hu2018squeeze, bello2019attention}, segmentation \cite{fu2019dual}, and object detection \cite{srinivas2021bottleneck}. Particularly in the detection of small objects, the technique is applied to guide the model's attention to very small details without incurring a considerable computational cost. Following the feature maps generated by convolutional layers, the attention masks are then computed to capture the spatial (Non-local Net \cite{wang2018non} and Criss-cross Net \cite{huang2019ccnet}) and channel-related information (SE \cite{hu2018squeeze} and AiA \cite{fang2021attention}). Also, some techniques such as CBAM \cite{woo2018cbam} and CA \cite{hou2021coordinate} combine the two for improved information fusion. In 3D vision tasks, Point-attention \cite{feng2020point} proposes a self-attention structure and skip-connection to capture long-range dependencies of points for point cloud segmentation. Point-Transformer \cite{zhao2020point} learns the attention weights for the neighbour points, by using local geometric relationships between the center point and its surrounding.

Following the success of pillar-based approaches \cite{lang2019pointpillars, liu2020tanet}, PiFeNet is a novel end-to-end trainable 3D object detector capable of extracting important features of small objects in real-time, which benefits from the reduced inference latency of single-stage methods and the improved feature extraction mechanisms from attention.

\section{Our Approach}
This section introduces pillar-based PiFeNet for 3D object detection, followed by the formulas of multi-point-aware pooling, channel-wise, point-wise, task-aware attention mechanisms in the PAA module, and the Mini-BiFPN's structure. 

\subsection{Pillar-Feature Network (PiFeNet)}

As shown in \cref{fig:PiFeNet}, PiFeNet firstly takes a point cloud as input and quantises it into pillars by a pillar generator module similar to \cite{lang2019pointpillars}. Next, we stack our PAA modules to extract pillar features with reduced information loss. Then the attention-weighted pillar features are scattered onto a pseudo-image, which is run through our Mini-BiFPN to fuse features on different resolutions. Finally, we use a simple detection head including classification and box regression branches to produce detection results.

\subsection{3D object detection} \label{sec:objectdetectiondefinition}
3D object detection expects to predict a 3D bounding box ($c_x, c_y, c_z, w, l, h, \theta)$ for each object. $c_x, c_y, c_z$ denote the x, y, and z centers of the box, respectively. $w, l, h$ are the box's dimensions and $\theta$ is the heading angle. The input of the model is a set of points (point cloud), defined as $T=\left\{p_{i}=\left[x_{i}, y_{i}, z_{i}, r_{i}\right]^{\top} \in \mathbb{R}\right\}_{i=1,2,3, \ldots, M}$. $x_{i}, y_{i}, z_{i}$ are the x, y, and z coordinates of points, respectively. $r_{i}$ is the points' reflection intensity, and $M$ is the total number of points. Assume the dimension of a point cloud $T$ is defined as $L^*$ for length ($x$ axis), $W^*$ for width (y-axis), and $H^*$ for height (z-axis). $T$ is then equally divided into pillars \cite{lang2019pointpillars} (z-axis stacked voxels) along the x and y axes. Due to the sparsity and unequal distribution of the point clouds, each pillar has a varying amount of points. Let $N$ be a pillar's maximum number of points, $C$ be the number of channels of each point, and $P$ be the amount of pillars in the pillar grid $G$, where $G=\left\{p_{1}, p_{2}, \ldots p_{j}\right\}_{j \leq P}$. Each pillar's dimensions are $[p_{W}^{*}, p_{L}^{*}, p_{H}^{*}]$, where $p_{H}^{*} = H^*$ since we do not divide the point cloud along z-axis. Let the number of output classes be $N_c$, and $N_a$ is the number of anchors per pillar. Our PiFeNet directly takes a raw point cloud as input and produces $P\times N_a \times N_c$ prediction boxes and their classifications. 

\subsection{Pillar Aware Attention (PAA) module}
\begin{figure}[tb]
\centering
\includegraphics[width=\linewidth]{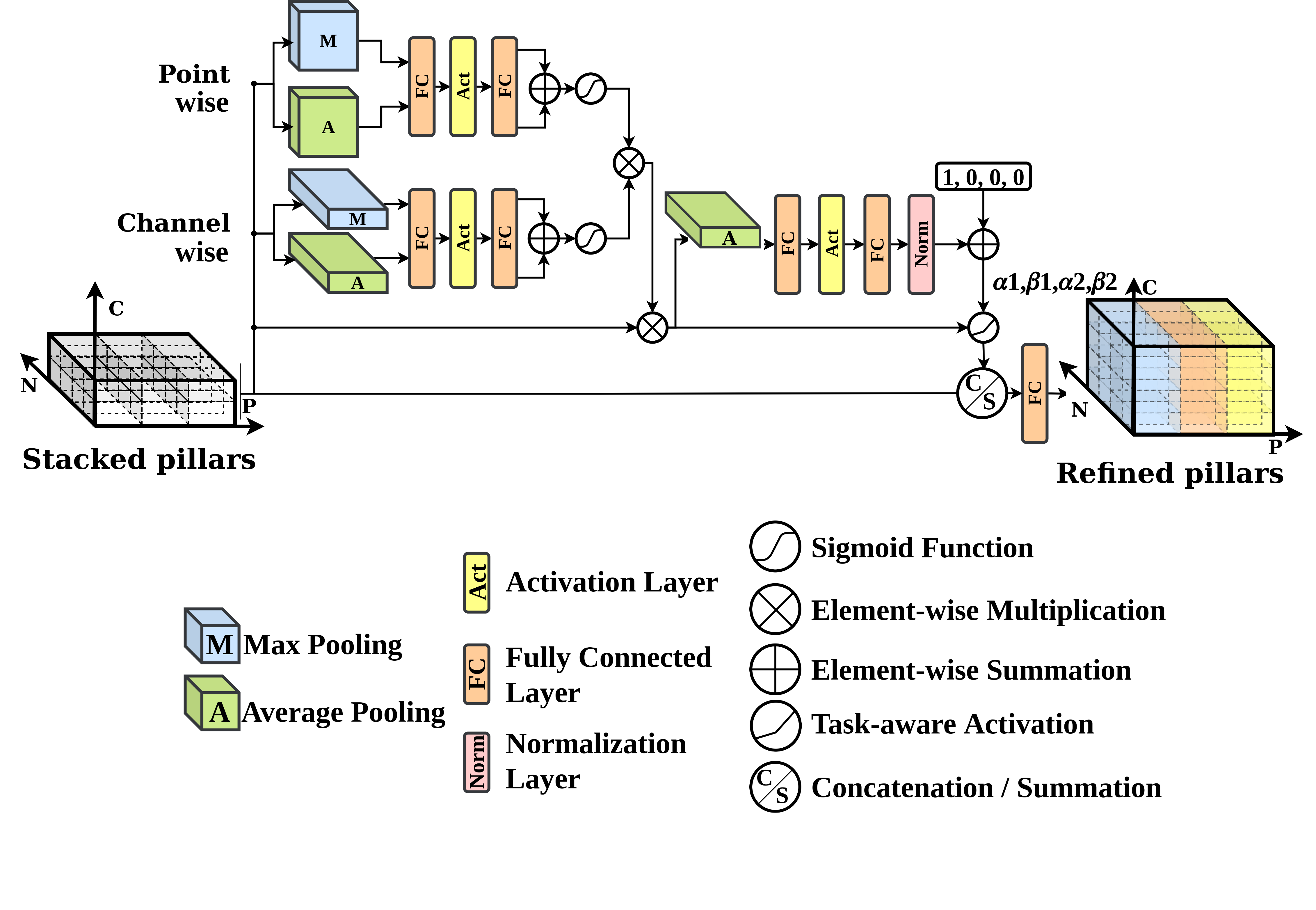}
\vspace{-1.5em}
   \caption{\tho{Detailed architecture of the Pillar Aware Attention module.}}
   \vspace{-1.5em}
\label{fig:PAAmodule}
\end{figure}

As shown in \cref{fig:PAAmodule}, our PAA module begins with multi-point-channel pooling followed by two branches of point- and channel-wise attention. The outputs are then integrated by element-wise multiplication before going through the task-aware attention sub-module. 

\textbf{Multi-point-channel pooling.} To capture the context of all points and channels in a pillar, we present a multi-point-channel pooling, which applies both max and average pooling to the point- and channel-wise dimensions in point clouds. Consider a pillar grid $G\in \mathbb{R}^{P\times N\times C} $, where $N$ is the maximum number of points, $C$ is the number of channels in a pillar, and $P$ is the maximum amount of pillars $\left\{p_{1}, p_{2}, \ldots p_{j}\right\}_{j \leq P}$ in the grid. \tho{We perform channel-wise average and max pooling on the pillar grid $G$ to aggregate channel information}. The outputs are two distinct channel context representations $F_c^{mean}$ and $F_c^{max}$, where $F_c^{max}, F_c^{mean} \in \mathbb{R}^{P\times 1\times C}$. The same strategies are also applied to the point-wise dimension to aggregate the point information $F_p^{mean}$ and $F_p^{max}$, where $F_p^{max}, F_p^{mean} \in \mathbb{R}^{P\times N\times 1}$. 

\textbf{Channel- and Point-wise attention.}  
To encapsulate the global information, $F_c^{mean}$ and $F_c^{max}$ are passed through a shared \tho{multi-layer perceptron (MLP) with two fully-connected layers, an activation function $\delta$, and a reduction ratio $r$ (the same applied to $F_p^{mean}$ and $F_p^{max}$). The outputs of the MLP are summed element-wise to generate the final attention score vectors $A_c \in \mathbb{R}^{P\times 1\times C}$ and $A_p\in \mathbb{R}^{P\times N\times 1}$:
\begin{equation}
    A_{c/p}=\sigma\left(w_{1}\left(w_{0}\delta\left(F_{c/p}^{mean}\right)\right) + w_{1}\left(w_{0}\delta\left(F_{c/p}^{max}\right)\right)\right)
    \label{eq:channel_att}
\end{equation}}
where $\sigma$ is Sigmoid function. The weights in the two fully-connected layers are $w_0 \in \mathbb{R}^{N/r\times N}$ and $w_1 \in \mathbb{R}^{N\times N/r}$ in case of point-wise attention, or $w_0 \in \mathbb{R}^{C/r\times C}$ and $w_1 \in \mathbb{R}^{C\times C/r}$ for channel-wise attention.
The full attention matrix is then achieved by combining channel-wise attention $A_c$ and point-wise attention $A_p$ using element-wise multiplication $M_j = A_p \times A_c$, where $M_j \in \mathbb{R}^{N\times P\times C}$. By multiplying $M_j$ with the original pillar $P_j$, we can obtain attention-weighted features in both channel-wise and point-wise dimensions.

After the point-wise and channel-wise attention sub-modules, the pillar features become more expressive and sensitive to all points in the point cloud and their channel features (x, y, z locations, centres).

\textbf{Task-aware attention.}
The task-aware attention sub-module is chained at the end to dynamically control the activation of each channel. The output pillar features are then reorganised into distinct activations in response to the needs of various downstream tasks \tho{(\ie detection and classification). Similar to \cite{chen2020dynamic}, given the feature map $F_{c}\in R^{1\times N}$ of the pillar $p_j$ and the channel $C_k$, the activation is calculated as follows:}
\begin{equation}
    A_{T}^{k}(F_{c})  =\max\left(\alpha^{1}_k\cdot F_{c}+\beta^{1}_k, \alpha^{2}_k\cdot F_{c}+\beta^{2}_k\right)
    \label{eq:task_att}
\end{equation}
\tho{
where $A_{T}^{k}(F_{c})$ is the task-aware weighted pillar feature map of the channel $C_k$. $[\alpha^1_k, \beta^1_k, \alpha^2_k, \beta^2_k]=\theta(\cdot)$ is a hyper-function to control channel-wise activation thresholds. It is constructed from one global pooling, two fully-connected layers and a shifted sigmoid layer $f(x) = 2 \sigma(x)-1$ to normalise the output to $[-1,1]$ before multiplying with the original features. Note that $\theta(\cdot)$ is initialised with $[1,0,0,0]$, indicating that \cref{eq:task_att} is originally a ReLU function $\textit{max}\{F_c,0\}$. Finally, the task-aware weighted features are concatenated or summed back to the input features.}

\textbf{Stackable PAA module.} Our approach stacks two PAA modules to better leverage multi-level feature attention. \tho{The first module combines nine task-aware weighted features with the original ones. The resulting 18 channels are then used as input to the second PAA module. The PAA results are element-wise added to the input of a fully-connected layer. The fully-connected layer increases the feature dimension to 64 for greater expressiveness}. Eventually, a point-wise max pooling is performed to extract the strongest features, serving as inputs of Mini-BiFPN module, as shown in \cref{fig:PiFeNet}.


\subsection{Mini-BiFPN}

\tho{Given the small occupancy of pedestrians in point clouds, the feature network needs more comprehensive connections, so that features from multiple resolutions can complement and compensate each other. Therefore, we propose Mini-BiFPN detection head, a faster variant of BiFPN \cite{tan2020efficientdet} that significantly improves the performance in 3D object detection with minimal efficiency trade-offs.} Mini-BiFPN firstly inputs a pseudo-image into convolutional blocks $B_{1},B_{2},B_{3}$ to produce a list of features $\bar{F}_{\text {in }}=\left(F_{in}^{1}, F_{in}^{2}, F_{in}^{3}\right)$, with resolutions $1/2^{i-1}$ of input pseudo-image where $i \in(1,2,3)$. Next, the multi-scale features are aggregated by repeatedly applying top-down and bottom-up bidirectional feature fusions, as shown in \cref{fig:PiFeNet}. Trainable weights are added to adjust the fusion weights accordingly. Our Mini-BiFPN can be formalised as follows:


\begin{equation}
\scalebox{0.9}{%
$F_{u p}^{2}=\operatorname{conv}\left(\delta\left(\frac{\left(w_{1}^{\prime} \cdot F_{i n}^{2}+w_{2}^{\prime} \cdot upsample\left(F_{i n}^{3}\right)\right.}{w_{1}^{\prime}+w_{2}^{\prime}+\varepsilon}\right)\right)$}
    \label{eq:f2_up}
\end{equation}
\begin{equation}
    \scalebox{0.9}{%
$F_{out}^{2}\!=\!\operatorname{conv}\!\left(\!\delta\!\!\left(\!\frac{\left(w_{1}^{\prime\prime} \cdot F_{in}^{2} +w_{2}^{\prime\prime} \cdot F_{up}^{2}+ w_{3}^{\prime\prime} \cdot downsample\left(F_{out}^{1}\right) \right.}{w_{1}^{\prime\prime}+w_{2}^{\prime\prime}+w_{3}^{\prime\prime} +\varepsilon}\!\right)\!\!\right)$}
    \label{eq:f2}
\end{equation}
where \tho{$\delta$ is Swish activation function adopted from \cite{tan2020efficientdet}}, and $F^2_{up}$ is the fusion result of $F^2_{in}$ and $F^3_{in}$. Then $F_{out}^{2}$ is calculated by $F^2_{up}$, $F_{out}^{1}$, and $F_{in}^{2}$. $w_{1,..,i}^{\prime}$ and $w_{1,..,i}^{\prime\prime}$ are trainable parameters where $i$ is the nu~\cite{ramachandran2017searching}. $F_{out}^{1}$ and $F_{out}^{3}$ are computed similarly to $F_{out}^{2}$. The final feature representation is the concatenation of $F_{out}^{1}$, $F_{out}^{2}$, and $F_{out}^{3}$. It is then passed into a simple SSD\cite{liu2016ssd} detection head including classification and box regression branches for final predictions.

\tho{\textbf{Loss function.} Similar to SECOND~\cite{yan2018second}, Focal loss~\cite{lin2017focal} and L1 loss are used for bounding box classification and regression, respectively. }

\section{Experiments} \label{sec:Experiments}

\begin{table*}[!htp]\centering
\scriptsize
\begin{tabular}{clc|cccc|cccc|cc}\toprule
&&&\multicolumn{8}{c|}{\textbf{Performance (mAP)}}&\multicolumn{2}{c}{\textbf{Speed \& Computation}}
\\\cmidrule{4-11}\cmidrule{12-13}
\multirow{2}{*}{\textbf{Stage}} &\multirow{2}{*}{\textbf{Method}} &\multirow{2}{*}{\textbf{Venue}} &\multicolumn{4}{c|}{\textbf{Birds-eye-view}} &\multicolumn{4}{c|}{\textbf{3D Detection}} & \multirow{2}{*}{\textbf{fps}}&\multirow{2}{*}{\textbf{GPU}} \\\cmidrule{4-7}\cmidrule{8-11}
& & &\textbf{Easy } &\textbf{Mod.} &\textbf{Hard} & \textbf{Overall}  &\textbf{Easy } &\textbf{Mod.} &\textbf{Hard} &\textbf{Overall}  &  \\\cmidrule{1-13}
\multirow{9}{*}{Two} &\textbf{PointPainting\cite{vora2020pointpainting}} &CVPR'20 &58.7 &49.93 &46.29 &51.64 &50.32 &40.97 &37.87 &43.05 &3* &Unknown \\\cmidrule{2-13}
&\textbf{AVOD-FPN\cite{ku2018joint}} &IROS'18 &58.49 &50.32 &46.98 &51.93 &50.46 &42.27 &39.04 &43.92 &10* &Titan X\\\cmidrule{2-13}
&\textbf{STD\cite{yang2019std}} &ICCV'19 &60.02 &48.72 &44.55 &51.10 &53.29 &42.47 &38.35 &44.70 &13* & Titan V\\\cmidrule{2-13}
&\textbf{F-PointPillars\cite{paigwar2021frustum}} &ICCV'21 &60.98 &52.23 &48.3 &53.84 &51.22 &42.89 &39.28 &44.46 &17*& 1080\\\cmidrule{2-13}
&\textbf{VMVS\cite{ku2019improving}} &IROS'19 &60.34 &50.34 &46.45 &52.38 &53.44 &43.27 &39.51 &45.41 &4* & Titan X\\\cmidrule{2-13}
&\textbf{VoxelToPoint\cite{li2021voxel}} &ACMMM'21 &56.54 &48.15 &45.63 &50.11 &51.8 &43.28 &40.71 &45.26 &10 &V100\\\cmidrule{2-13}
&\textbf{PV-RCNN\cite{shi2020pv}} &CVPR'20 &59.86 &50.57 &46.74 &52.39 &52.17 &43.29 &40.29 &45.25 &13*& 1080\\\cmidrule{2-13}
&\textbf{PartA2\cite{shi2020points}} &TPAMI'20 &59.04 &49.81 &45.92 &51.59 &53.1 &43.35 &40.06 &45.50 &13&V100 \\\cmidrule{2-13}
&\textbf{F-ConvNet\cite{wang2019frustum}} &IROS'19 &57.04 &48.96 &44.33 &50.11 &52.16 &43.38 &38.8 &44.78 &2* &Unknown\\\cmidrule{2-13}
&\tho{\textbf{EQ-PVRCNN\cite{EQPVRCNN}}} &CVPR'22 &61.73 &52.81 &49.87 &54.80 &55.84 &\textbf{47.02} &\textbf{42.94} &\textbf{48.60} &5&3090 \\\midrule
\multirow{9}{*}{One} &\textbf{VoxelNet} &CVPR'18 &57.73 &39.48 &33.69 &43.63 &39.48 &33.69 &31.51 &34.89 &4* &Unknown\\\cmidrule{2-13}
&\textbf{SECOND\cite{yan2018second}} &Sensors'18 &55.1 &46.27 &44.76 &48.71 &51.07 &42.56 &37.29 &43.64 &20* &1080\\\cmidrule{2-13}
&\textbf{PointPillars\cite{lang2019pointpillars}} &CVPR'19 &57.6 &48.64 &45.78 &50.67 &51.45 &41.92 &38.89 &44.09 &63&V100 \\\cmidrule{2-13}
&\textbf{MGAFSSD\cite{li2021anchor}} &ACMMM'21 &56.09 &48.46 &44.9 &49.82 &50.65 &43.09 &39.65 &44.46 &10 &V100\\\cmidrule{2-13}
&\textbf{Point-GNN\cite{shi2020point}} &CVPR'20 &55.36 &47.07 &44.61 &49.01 &51.92 &43.77 &40.14 &45.28 &2 &V100\\\cmidrule{2-13}
&\textbf{3DSSD\cite{yang20203dssd}} &CVPR'20 &60.54 &49.94 &45.73 &52.07 &54.64 &44.27 &40.23 &46.38 &25* &Titan V\\\cmidrule{2-13}
&\textbf{TANet\cite{liu2020tanet}} &AAAI'20 &60.85 &51.38 &47.54 &53.26 &53.72 &44.34 &40.49 &46.18 &29* &Titan V\\\cmidrule{2-13}
&\textbf{HotSpotNet\cite{chen2020object}} &ECCV'20 &57.39 &50.53 &46.65 &51.52 &53.1 &45.37 &41.47 &46.65 &25&V100 \\\cmidrule{2-13}
&\textbf{PiFeNet (ours)} & &\textbf{63.25} &\textbf{53.92} &\textbf{50.53} &\textbf{55.90} &\textbf{56.39} &46.71 &42.71 &\textbf{48.60} &26&V100 \\
\cmidrule{1-13}
\end{tabular}
\caption{Comparison with previous approaches for 3D object detection task on the KITTI official \textit{test} split for \textit{Pedestrian} on birds-eye-view (BEV) and 3D detection tasks. The AP is calculated using 40 recall positions, and 3D mAP represents the mean average precision of the three difficulties. ``*" denotes run-time is cited from KITTI official website.}\vspace{-2em}
\label{table:pedestriancomparision}
\end{table*}   

Our proposed PiFeNet is evaluated on KITTI \cite{geiger2013vision}, JRDB \cite{martin2021jrdb} and Nuscenes \cite{caesar2020nuscenes} benchmarks. We choose CBGS-PP (Pointpillars backbone) as a baseline for Nuscenes and PointPillars for KITTI and JRDB. \tho{We tried training PiFeNet on KITTI using pretrained weights on JRDB and observed that the pretraining-finetuning strategy is not necessary considering the domain shift, sensor setup, and LiDAR resolution differences among datasets. Similar behaviours are also seen in \cite{jia2022domain}. Therefore, all models in this work are trained from scratch (random weight initialisation). }

\textbf{KITTI} has 7481 and 7518 samples for training and testing, respectively. The training set is further split into training and validating set with a ratio of 85:15. We conduct experiments on the "Pedestrian" category and use average precision (AP) with an (IoU) threshold of 0.5 as an evaluation metric. The detection range is [0, 47.36], [-19.84, 19.84], [-2.5, 0.5] metres in X and Y and Z coordinates, respectively. Pillar size is [0.16, 0.16, 4] metres in width, length, and height. Also, ground truths are mixed with sampled objects from other point clouds, together with random global rotation, scaling, and translation.

\textbf{JRDB} is a novel human-centric dataset comprised of more than 1.8 million 3D cuboids surrounding all pedestrians in both indoor and outdoor settings. This dataset contains 54 sequences, 27 for training and 27 for testing. During training, we select 5 from 27 sequences for validation, as suggested in \cite{martin2021jrdb}, and we use AP with an IoU of 0.5 to monitor the training process. The detection range is [-39.68, 39.68] in X and Y, and [-2, 2] in Z LiDAR coordinate. Random pedestrians are cut-and-pasted to the scene ensuring at least 30 people per input point cloud. Random rotation, flipping and scaling are also applied. The top 300 proposals after Non-max suppression (NMS) with IoU of 0.5 are kept as final predictions.

\textbf{Nuscene.} We adopt CBGS \cite{zhu2019class} augmentations strategies (global translation, scaling, and rotation) and multi-task heads. The detection range is [-51.2, 51.2] metres for X and Y, and [-5, 3] for Z LiDAR coordinates, pillar size is set at [0.2, 0.2, 8] in width, length, and height respectively. We train the model for 20 epochs with a batch size of 2. During testing, top 83 predictions with confidence scores greater than 0.1 are kept, and rotated NMS with IOU threshold of 0.02 is applied.

AdamW \cite{loshchilov2017decoupled} optimiser and One Cycle policy \cite{smith2019super} with max LR at 0.003 are used in all experiments. 

\subsection{Comparison with State-of-the-arts}

\label{sec:ComparisionSOTA}
\begin{figure*}[!th]
\begin{center}
\includegraphics[width=\linewidth]{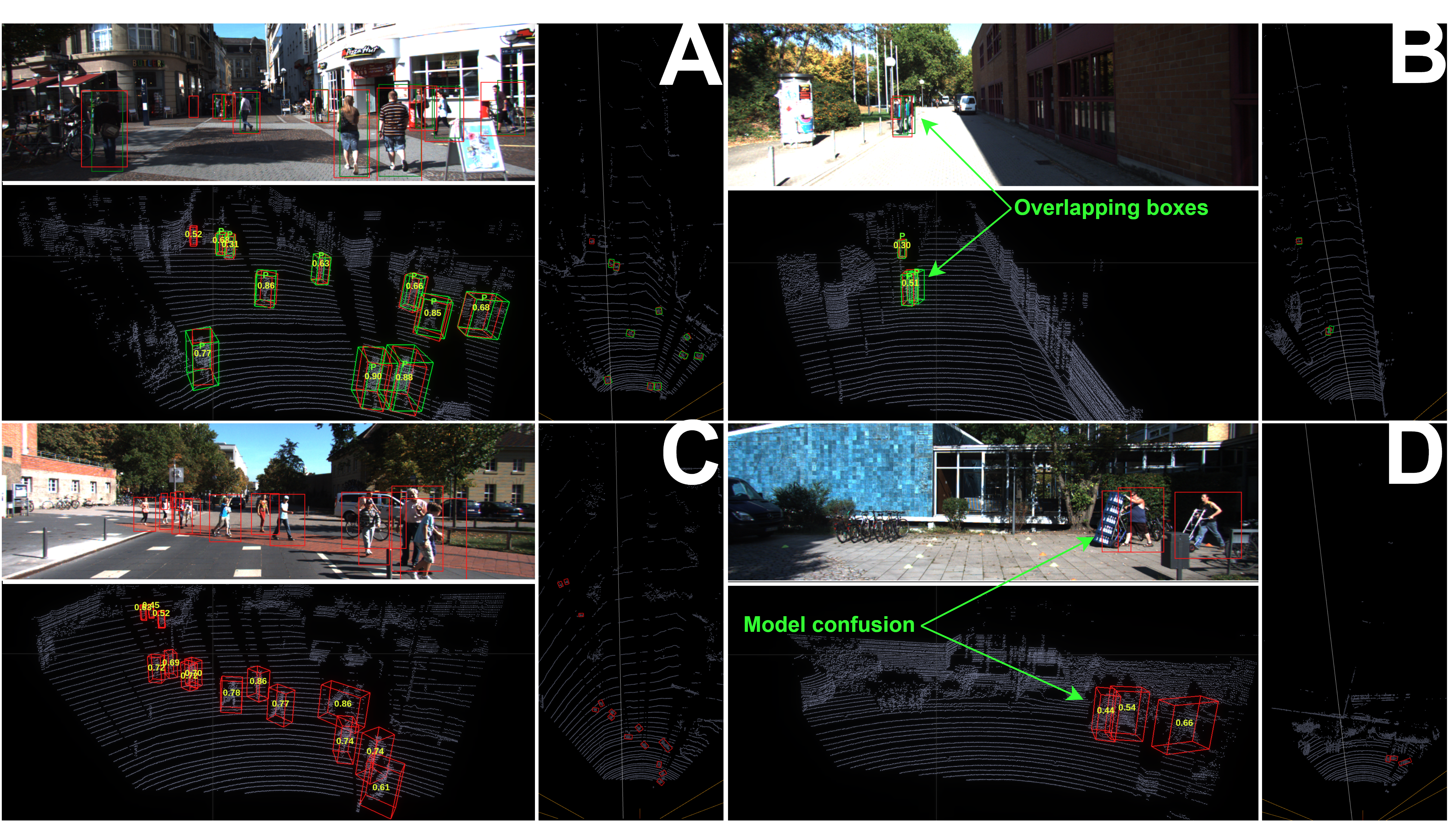}
\end{center}
\vspace{-1em}
   \caption{Visualization on KITTI val set (A, B) and test set (C, D). Each visualization includes BEV (right), 3D point clouds (top), and 2D image (bottom). The ground truth boxes are green, the predicted boxes are red with yellow confident scores. B and D are typical val and test set failures.}
   \vspace{-1em}
\label{fig:PiFeNetviz}
\end{figure*}
\textbf{KITTI.} As shown in \cref{table:pedestriancomparision}, PiFeNet outperforms all the existing state-of-the-art approaches in BEV detection. Our model outperforms TANet \cite{liu2020tanet} by 2.64 mAP in BEV and 2.22 mAP in 3D detection. It also surpasses the current strongest BEV pillar-based detector Frustum-PointPillars \cite{paigwar2021frustum} by 2.06 mAP in BEV pedestrian detection\footnote{\textit{cvlibs.net/datasets/kitti/eval\_object.php?obj\_benchmark=bev}}. These significant improvements are caused by the enhanced expressiveness of the pillars, together with better communication between pillars at multiple scales. While achieving SOTA performance, PiFeNet's inference speed is 26 FPS (\cf \cref{table:modules_ablation}), making it one of the most efficient detectors on KITTI.

\begin{table}[tb]\centering
\scriptsize
\begin{tabular}{lcccc}\toprule
\textbf{Method} &\textbf{AP (pedestrian)} &\textbf{mAP (all classes)} &\textbf{NDS} \\\cmidrule{1-4}
\textbf{PointPillars+\cite{lang2019pointpillars}} &0.640 &0.401 &0.550 \\\cmidrule{1-4}
\textbf{WYSIWYG\cite{WYSIWYG}} &0.650 &0.350 &0.419 \\\cmidrule{1-4}
\textbf{SARPNET\cite{ye2020sarpnet}} &0.694 &0.324 &0.484 \\\cmidrule{1-4}
\textbf{SA-Det3D\cite{sadet3d}} &0.733 &0.470 &0.592 \\\cmidrule{1-4}
\textbf{PiFeNet (ours)} &\textbf{0.745} &\textbf{0.478} &\textbf{0.606} \\\cmidrule{1-4}
\end{tabular}
\caption{PiFeNet performance on official Nuscenes test set compared to other approaches. Bold values are the best performance. }\label{tab:nuscenesresult}\vspace{-2em}
\end{table}
\textbf{JRDB and nuScenes.} We further evaluate our model on JRDB-3D detection benchmark. Compared to other published methods, our model is ranked first on JRDB 2019\footnote{\textit{https://jrdb.erc.monash.edu/leaderboards/detection}} in multiple AP thresholds (\ie 0.3, 0.5, and 0.7). Our model also achieves the best performance on the recently launched JRDB 2022\footnote{\textit{https://jrdb.erc.monash.edu/leaderboards/detection22}} 3D detection leaderboard. We also present in \cref{tab:nuscenesresult} the performance of our approach on nuScenes. Compared with our baseline Pointpillars++\cite{lang2019pointpillars}, there is a significant improvement of 0.11 AP in pedestrian detection. 
\begin{table}[tb]\centering
\scriptsize
\begin{tabular}{cccccc}\toprule
&\textbf{Method} &\textbf{AP@0.3} &\textbf{AP@0.5} &\textbf{AP@0.7} \\\cmidrule{1-5}
\multirow{4}{*}{JRDB 2019} &F-Pointnet\cite{qi2018frustum} &38.205 &6.378 &0.081 \\\cmidrule{2-5}
&EPNet\cite{huang2020epnet} &59.252 &16.845 &0.418 \\\cmidrule{2-5}
&TANet++\cite{liu2020tanet} &63.922 &27.991 &1.842 \\\cmidrule{2-5}
&PiFeNet (ours) &\textbf{74.284} &\textbf{42.617} &\textbf{4.886} \\\midrule
\multirow{2}{*}{JRDB 2022} &PointPillars*\cite{lang2019pointpillars} &69.209 &33.677 &2.209 \\\cmidrule{2-5}
&PiFeNet (ours) &\textbf{70.724} &\textbf{39.838} &\textbf{4.59} \\
\cmidrule{1-5}
\end{tabular}
\caption{PiFeNet performance on official the JRDB 2019 and newly launched JRDB 2022 test sets compared to other approaches. *reproduced result}\label{tab:jrdb} \vspace{-3em}
\end{table}

\begin{figure*}[!ht]
\vspace{-1em}
\begin{center}
\includegraphics[width=0.99\linewidth,height=6.3cm]{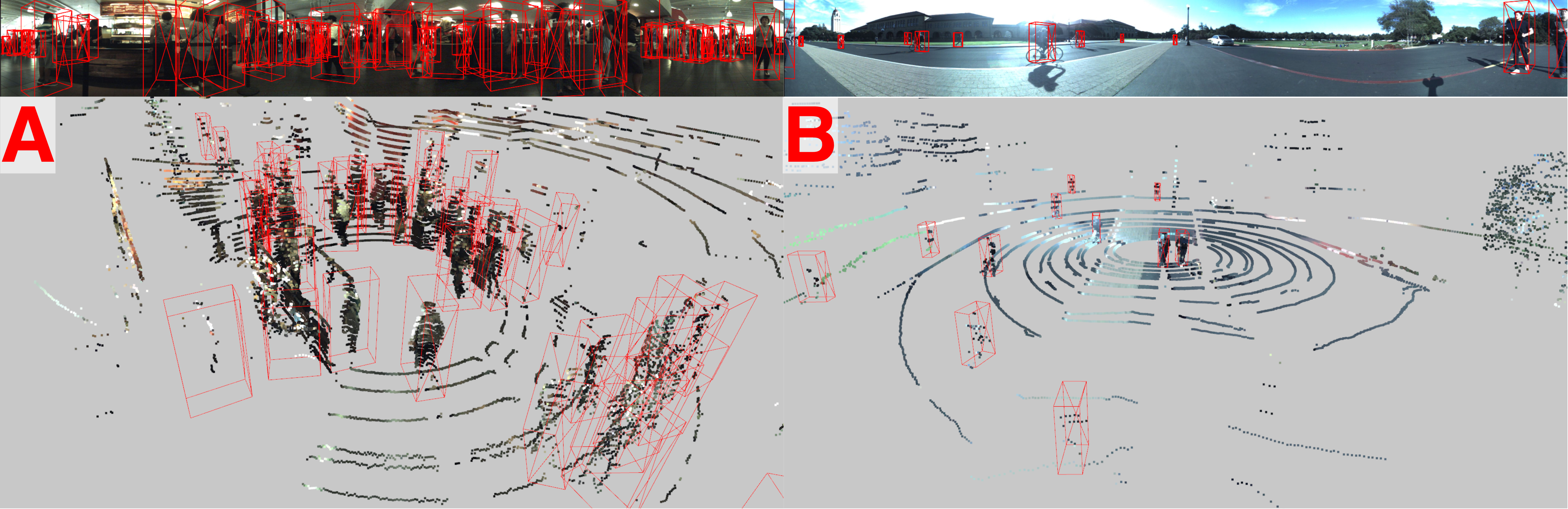}
\end{center}
\vspace{-1em}
   \caption{Qualitative results on JRDB 2022 official test set. Very crowded indoor (A) and outdoor (B) settings are visualised (2D panorama image at the top, 3D point cloud at the bottom). Prediction bounding boxes (with red cross showing heading direction) are shown in red. Best viewed in color.}
  \vspace{-1.5em}
\label{fig:JRDBviz}
\end{figure*}

In \cref{fig:PiFeNetviz}, we present our qualitative findings which include both detection results (\cref{fig:PiFeNetviz}A and \ref{fig:PiFeNetviz}C) as well as some typical failure cases (\cref{fig:PiFeNetviz}B and \ref{fig:PiFeNetviz}D). \cref{fig:PiFeNetviz}C shows the robustness of our model when it can accurately detect all pedestrians in the picture, even at very long distances. Compared to Pointpillars \cite{lang2019pointpillars}, we successfully overcome the lack of expressiveness in the extracted pillar features, which usually leads to confusion between narrow vertical features of trees/poles and pedestrians. Given the fact that we rarely see the model misclassify a tree/pole as a pedestrian, we did find an instance where the model was confused when encountering a stack of soda crates as in \cref{fig:PiFeNetviz}D. The second difficulty arises when pedestrians are too close together (\cref{fig:PiFeNetviz}B), preventing the model from accurately locating both items. 
However, as shown in \cref{fig:JRDBviz}A, PiFeNet can elegantly distinguish people from background objects in a very crowded indoor scene, thanks to the improved representation and multi-scale communication between pillars.

\section{Ablation Studies} \label{sec:AblationStudy}
We provide extensive ablation experiments to evaluate the efficacy and contribution of each component in PiFeNet. These experiments are conducted on the official KITTI \textit{val} split. We use Pointpillars \cite{lang2019pointpillars} as our baseline. 
Our baseline without any attention mechanisms reaches 61.43 overall mAP, and achieves 66.73, 61.06, and 56.5 AP at three difficulty levels (easy, moderate, and hard).

\begin{table}[h]\centering
\scriptsize
\begin{tabular}{cc|ccccccccc}\toprule
\multicolumn{2}{c}{\textbf{Pooling }} &\multicolumn{3}{c}{\textbf{Modules}} &\multicolumn{5}{c}{\textbf{Pedestrian}} \\\cmidrule{1-10}
\rotatebox{90}{\textbf{Avg}} &\rotatebox{90}{\textbf{Max}} &\rotatebox{90}{\textbf{Point-}} &\rotatebox{90}{\textbf{Channel-}} &\rotatebox{90}{\textbf{Task-}} &\rotatebox{90}{\textbf{Easy}} &\rotatebox{90}{\textbf{Moderate}} &\rotatebox{90}{\textbf{Hard}} &\rotatebox{90}{\textbf{mAP}} &\rotatebox{90}{\textbf{FPS}} \\\cmidrule{1-10}& & & & &66.7 &61.1 &56.5 &61.4 &41.5 \\
\cmark &\cmark &\cmark & & &69.6 &63.4 &58.1 &63.7 &36.3 \\
\cmark &\cmark & &\cmark & &67.3 &62.1 &56.4 &61.9 &35.0 \\
\cmark &\cmark & & &\cmark &68.2 &61.4 &55.9 &61.8 &31.2 \\
\cmark &\cmark &\cmark &\cmark & &67.3 &61.4 &56.5 &61.8 &32.3 \\
\cmark &\cmark & &\cmark &\cmark &69.3 &63.3 &58.1 &63.6 &28.1 \\
\cmark &\cmark &\cmark & &\cmark &68.3 &62.4 &56.6 &62.4 &28.8 \\\cmidrule{1-10}
\cmark & &\cmark &\cmark &\cmark &68.2 &62.2 &57.3 &62.6 &28.8 \\
&\cmark &\cmark &\cmark &\cmark &69.6 &61.9 &55.4 &62.3 &27.9 \\
\cmark &\cmark &\cmark &\cmark &\cmark &\textbf{69.7} &\textbf{64.0} &\textbf{58.7} &\textbf{64.2} &27.4 \\
\cmidrule{1-10}
\end{tabular}
\caption{Ablation study on PAA module's components.}\label{tab:ablationstudy}\vspace{-3em}
\end{table}

\textbf{Attention module analysis.} We use both average and max pooling as a default setting, then we sequentially test each of the PAA module's components and their combinations. According to~\cref{tab:ablationstudy}, channel-wise and task-aware attention alone does not significantly improve the performance. However, combining them results in a significant 2.2\% mAP increase. Moreover, integrating point-wise attention with either channel-wise or task-aware slightly improves the performance. Therefore, while standalone point-wise attention (2.3\% mAP increase) can significantly increase model performance, channel-wise and task-aware sub-modules must be employed together for better accuracy. 

\begin{table}[!h]\centering
\scriptsize
\begin{tabular}{lcccccc}\toprule
\multirow{2}{*}{\textbf{Modules}} &\multicolumn{5}{c}{\textbf{Pedestrian}} \\\cmidrule{2-6}
&\textbf{Easy} &\textbf{Moderate} &\textbf{Hard} &\textbf{mAP} &\textbf{FPS} \\\cmidrule{1-6}
\textbf{Baseline} &66.7 &61.1 &56.5 &61.4 &41.5 \\
\textbf{+ 1 PAA} &68.29 &62.36 &56.64 &62.43 &34.2 \\
\textbf{+ 2 PAA} &69.70 &64.02 &58.74 &64.2 &27.4 \\
\textbf{+ 3 PAA} &\textbf{70.37} &\textbf{64.35} &\textbf{58.92} &\textbf{64.55} &19.3 \\\cmidrule{1-6}
\end{tabular}
\caption{Ablation study on the number of stacked PAA modules}\label{tab:stackPAA_analysis}
\vspace{-2em}
\end{table}

\textbf{Number of stacked PAA modules.} \tho{\cref{tab:stackPAA_analysis} shows how the number of PAA modules affects the performance. Firstly, more PAA modules improve the accuracy, which demonstrates the effectiveness of our PAA module. Nevertheless, when we stack three PAA modules or more, the performance improvement rate slightly decelerates and the model requires more time for running.  
To balance the accuracy and speed, we use two PAA modules in other experiments.
}

\textbf{Pooling mechanisms analysis.} In the last 3 rows of \cref{tab:ablationstudy}, we show that each pooling method improves the performance (point-, channel-, and task-aware sub-modules are included as default). The z-axis related channels may be used to extract the maximum height of points inside a pillar, while the x and y-axis related channels can be used to extract the mean position of points inside a pillar. While average pooling outperforms max pooling in detecting moderate and hard objects, max pooling is better in detecting easy ones. Concurrent usage of both pooling these methods increases mAP to 64.2, showing that both pooling mechanisms are essential. 

\begin{table}[h]\centering
\scriptsize
\begin{tabular}{lccccc}\toprule
&\multicolumn{4}{c}{\textbf{Pedestrian}} &\multirow{2}{*}{\textbf{FPS}} \\\cmidrule{2-5}
&\textbf{Easy} &\textbf{Moderate} &\textbf{Hard} &\textbf{mAP} & \\\cmidrule{1-6}
\textbf{Baseline} &66.73 &61.06 &56.5 &61.43 &41.5 \\\cmidrule{1-6}
\textbf{ + PAA } &69.70 &64.02 &58.74 &64.2 &27.4 \\\cmidrule{1-6}
\textbf{ + Mini-BiFPN} &68.52 &62.30 &57.19 &62.67 &36.2 \\\cmidrule{1-6}
\textbf{ + PAA + Mini-BiFPN}
&\textbf{73.78} &\textbf{67.69} &\textbf{61.51} &\textbf{67.66} &25.7 \\
\cmidrule{1-6}
\end{tabular}
\caption{Contribution of modules to PiFeNet}\label{table:modules_ablation}\vspace{-1.5em}
\end{table}

\begin{figure}[!h]
\begin{center}
\includegraphics[width=\linewidth]{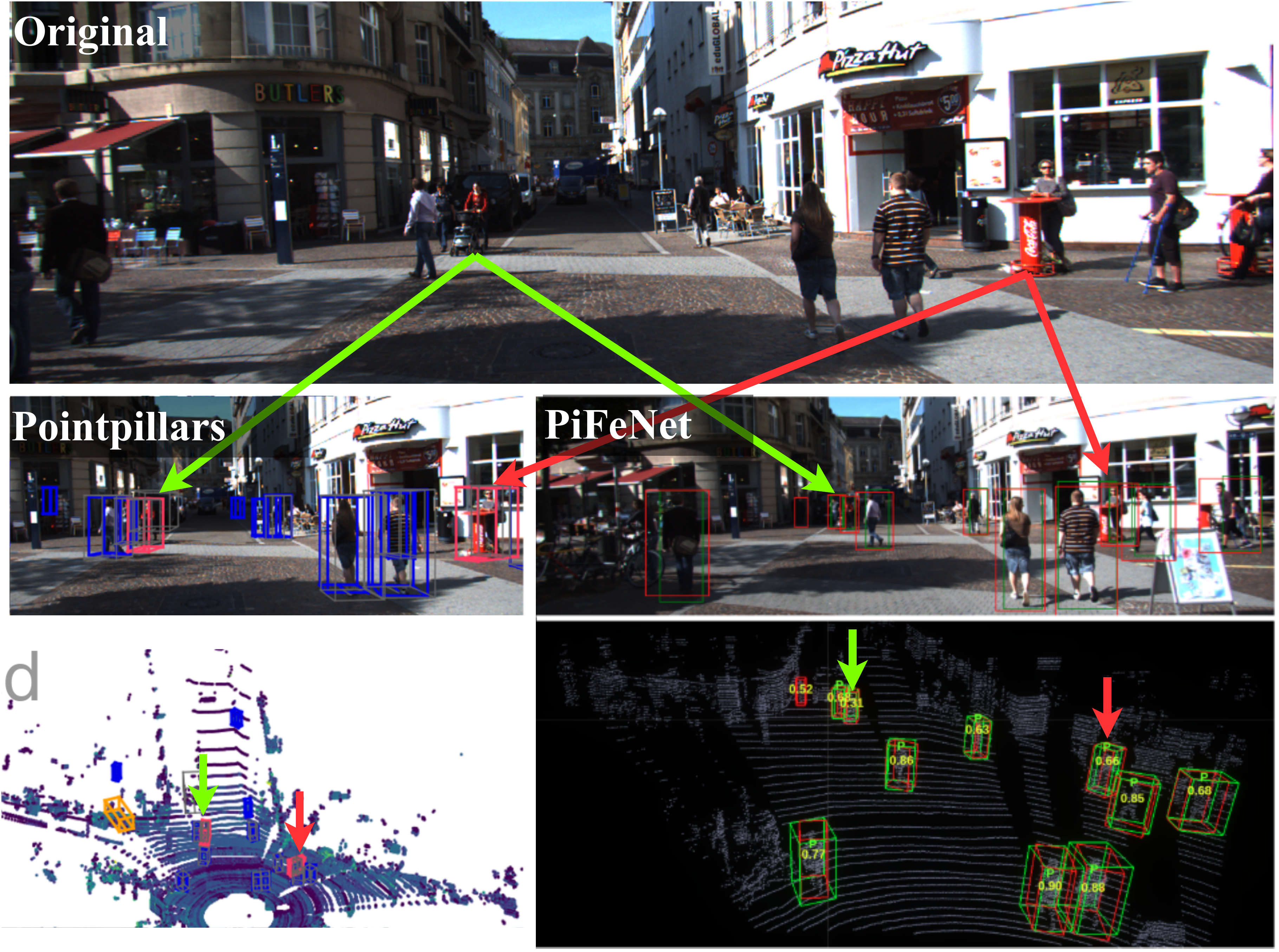}
\end{center}
\vspace{-1.05em}
   \caption{\tho{The images on the bottom left and right are taken from the PointPillars\cite{lang2019pointpillars} paper and our \cref{fig:PiFeNetviz}A, respectively. This scenario shows that the pedestrian standing next to the red pole and the one pushing a baby trolley are wrongly classified as a cyclists by PointPillars (baseline), whereas PiFeNet makes correct predictions. Best viewed in color.}}
   \vspace{-2em}
\label{fig:pointpillar_pifenet}
\end{figure}

\textbf{Mini-BiFPN modules analysis.} \cref{table:modules_ablation} also presents the outcome of the Mini-BiFPN module. 
\tho{By gathering more information from other stages of the feature network, our Mini-BiFPN's gate-controlled cross-scale connections and bidirectional information flow provide more accurate predictions.} As in \cref{table:modules_ablation}, after switching from the baseline feature network to the Mini-BiFPN module, we achieve a significant improvement in all APs (7.05, 6.63, and 5.01 increase in easy, moderate, and hard APs) as well as 6.23 increase in mAP compared to the baseline. Thus, both PAA and Mini-BiFPN modules are important components of our model. Additionally, the qualitative improvement over the PointPillars baseline is shown in \cref{fig:pointpillar_pifenet}.

\textbf{Run-time Analysis}
At inference time, PiFeNet can reach an average speed of 26 FPS (39.16 $ms$ latency), which breaks down into 5.84 $ms$ for data pre-processing, 18.12 $ms$ for PAA module, 0.60 $ms$ for scattering the features to a pseudo-image, 13.26 $ms$ for Mini-BiFPN, and 1.34 $ms$ for post-processing the predictions. The experiment is conducted on the KITTI dataset and a single NVIDIA V100 GPU. Following previous works, we also compare the inference speed (in FPS) between PiFeNet and other published approaches. While this comparison cannot be easily benchmarked due to hardware/computation variations, it still demonstrates the efficiency of our detector for running in real-time.

\section{Conclusion}
Our PiFeNet has been presented, addressing two main issues in 3D pedestrian detection. First, we introduced the Pillar Aware Attention Module, which combines multi-point-channel-pooling, point-wise, channel-wise, and task-aware attention to better extract pillar features. Next is the Mini-BiFPN module, a lightweight feature network that leverages cross-scale feature fusion and bidirectional connections, enriching information flow in the feature network. Our method achieves SOTA performance on both large-scale benchmarks KITTI \cite{geiger2013vision} and JRDB~\cite{martin2021jrdb}, and competitive results on Nuscenes \cite{caesar2020nuscenes}.

\bibliographystyle{splncs04}
\bibliography{egbib}

\vfill

\end{document}